\documentclass[letterpaper]{article} 
\usepackage[submission]{aaai23}  
\usepackage{times}  
\usepackage{helvet}  
\usepackage{courier}  
\usepackage[hyphens]{url}  
\usepackage{graphicx} 
\usepackage{natbib}  
\usepackage{caption} 
\usepackage{algorithm}
\usepackage{algorithmic}
\usepackage{amsmath}
\usepackage{newfloat}
\usepackage{listings}
\DeclareCaptionStyle{ruled}{labelfont=normalfont,labelsep=colon,strut=off} 
\floatstyle{ruled}
\newfloat{listing}{tb}{lst}{}
\floatname{listing}{Listing}
\usepackage{booktabs}
\usepackage{multirow}
\usepackage{color}

\title{A Novel Energy Based Model Mechanism for Multi-Modal Aspect-Based Sentiment Analysis}
\author{Tianshuo Peng$^{1,\dag}$, Zuchao Li$^{1,\dag,}$\thanks{$\ $  Corresponding author. $^\dag$ Equal contribution. This work was supported by the National Natural Science Foundation of China (No. 62306216), the Natural Science Foundation of Hubei Province of China (No. 2023AFB816), the Fundamental Research Funds for the Central Universities (No. 2042023kf0133), the Special Fund of Hubei Luojia Laboratory (No. 220100014), National Natural Science Foundation of China [No. 72074171] [No. 72374161].}, Ping Wang$^{3,4}$, Lefei Zhang$^{1,2}$, and Hai Zhao$^{5}$\\
$^{1}$School of Computer Science, Wuhan University, Wuhan, 430072, China \\
$^{2}$Hubei Luojia Laboratory, Wuhan 430072, P. R. China, \\
$^{3}$Center for the Studies of Information Resources, Wuhan University, Wuhan 430072, China\\
$^{4}$School of Information Management, Wuhan University, Wuhan 430072, China\\
$^{5}$Department of Computer Science and Engineering, Shanghai Jiao Tong University
{\tt \small \{pengts,zcli-charlie,wangping,zhanglefei\}@whu.edu.cn, zhaohai@cs.sjtu.edu.cn}\\
}

\usepackage{bibentry}

\begin{document}

\maketitle
\begin{abstract}
Multi-modal aspect-based sentiment analysis (MABSA) has recently attracted increasing attention. 
The span-based extraction methods, such as FSUIE, demonstrate strong performance in sentiment analysis due to their joint modeling of input sequences and target labels. However, previous methods still have certain limitations:
(i) They ignore the difference in the focus of visual information between different analysis targets (aspect or sentiment).
(ii) Combining features from uni-modal encoders directly may not be sufficient to eliminate the modal gap and can cause difficulties in capturing the image-text pairwise relevance.
(iii) Existing span-based methods for MABSA ignore the pairwise relevance of target span boundaries.
To tackle these limitations, we propose a novel framework called DQPSA for multi-modal sentiment analysis. Specifically, our model contains a Prompt as Dual Query (PDQ) module that uses the prompt as both a visual query and a language query to extract prompt-aware visual information and strengthen the pairwise relevance between visual information and the analysis target. Additionally, we introduce an Energy-based Pairwise Expert (EPE) module that models the boundaries pairing of the analysis target from the perspective of an Energy-based Model. This expert predicts aspect or sentiment span based on pairwise stability.
Experiments on three widely used benchmarks demonstrate that DQPSA outperforms previous approaches and achieves a new state-of-the-art performance. Furthermore, we conducted a fair comparison with relevant large-scale models such as ChatGPT-3.5 and VisualGLM. We found that our model has significant advantages in terms of performance though with relatively fewer parameters. A large amount of complementary experiments and ablation studies further demonstrate the effectiveness of the components we proposed.
The code will be released at~\url{https://github.com/pengts/DQPSA}.
\end{abstract}

\section{Introduction}

As one of the most important tasks that examines a model's semantic comprehension and sentiment perception, Multi-modal Aspect-Based Sentiment Analysis (MABSA) is an challenging and fine-grained task in the Sentiment Analysis field and has attracted increasing attention. The MABSA task consists of three main tasks: given an image-text pair, 
Multi-modal Aspect Term Extraction (MATE) focuses on extracting all aspect terms with sentiment polarity in the sentence~\cite{zhao-etal-2022-entity,lu-etal-2018-visual,DBLP:conf/nlpcc/WuCWLC20}, 
Multi-modal Aspect-oriented Sentiment Classification (MASC) aims to determine the sentiment polarity of each given aspect~\cite{DBLP:conf/aaai/XuMC19,DBLP:conf/ijcai/Yu019,DBLP:journals/taslp/YuJX20}, 
Joint Multi-modal Aspect-Sentiment Analysis (JMASA), on the other hand, requires the model to extract aspect-sentiment pairs jointly~\cite{ju-etal-2021-joint,ling-etal-2022-vision,zhou-etal-2023-aom}.
Among all the methods of previous work, the span-based extraction methods, such as FSUIE~\cite{peng-etal-2023-fsuie}, demonstrate strong performance in sentiment analysis due to their joint modeling of input sequences and target labels. Besides, it avoids complex structures for sequence labelling or sequence generation with a more concise structure.

\begin{figure}
    \centering    
    \includegraphics[width=1\columnwidth]{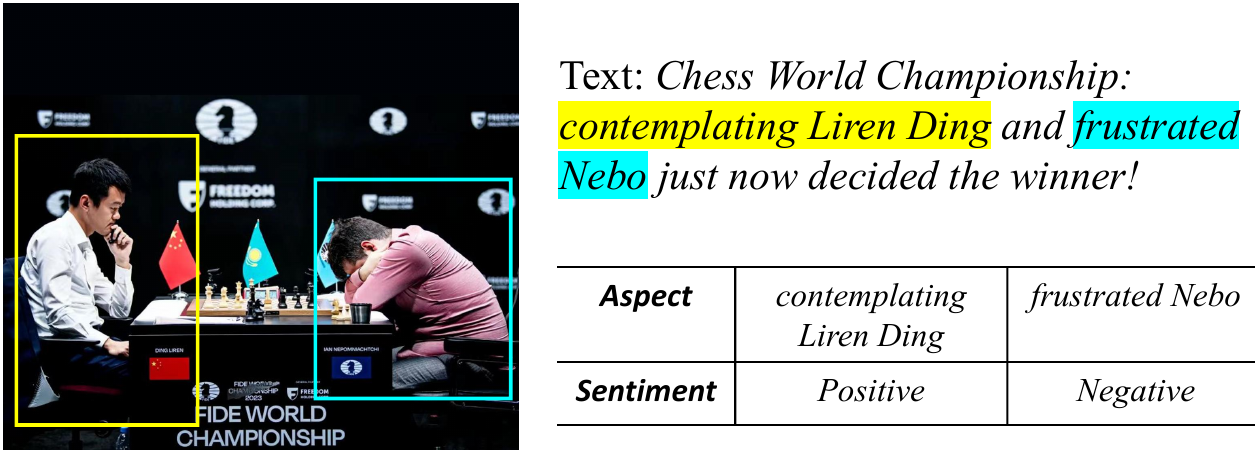}  
    \caption{Example of the variability in the focus of different analysis tasks}     
    \label{fig:focus_dif}    
\end{figure}

In this scenario of fine-grained MABSA task, three main challenges are worth emphasizing: 
First, image contains a large amount of semantic information, and there is a difference in the focus of visual information between different analysis targets. Take figure~\ref{fig:focus_dif} as an example: 
(1) The focus of visual information is different between MATE and MASC tasks: the MATE task should focus on all the potential entities across the image while MASC concentrates on the details of specific aspect which is fine-grained. (2) In the MASC task, different image regions may imply different sentimental tendencies, resulting the difference of focus among aspects.
Previous work only focused on the general information in image features, while the visual information representing positive emotions, such as the person pondering with his chin resting on their hand, and the visual information representing negative emotions, such as the person with his head lowered would influence each other, introducing a significant amount of misleading information into their sentiment analysis.

Second, most of the aforementioned studies focus on extracting modal features using pre-trained unimodal models and fusing them directly. However, separate training of image feature and text feature ignores the semantic alignment and modal gap between text and image, leading to the difficulty of the model in capturing the pairwise relevance between image and text. Therefore, it is crucial to design specific structures to mitigate modal gap and strengthen image-text pairwise relevance

Besides, existing span-based models~\cite{peng-etal-2023-fsuie} consider independently the possibility of certain position as a start or end boundary while ignoring the pairwise relevance between span boundaries, i.e., the a priori knowledge that "the boundaries of spans should be semantically related".

Based on the challenges above, we proposed the DQPSA framework for Multi-modal Aspect-Based Sentiment Analysis, which unifies the MABSA tasks under one framework. 
Specifically, inspired by BLIP-2~\cite{DBLP:journals/corr/abs-2301-12597} that trains an adapter to filter features from visual encoder,
we designed the \textit{Prompt as Dual Query} module to address the issue that different analysis targets pay different attention to visual information and strengthen the pairwise relevance between image and text.  
\textit{Prompt as Dual Query} uses prompt as both visual query and language query. The visual query interacts with image features from frozen pre-trained image encoder in alternating self-attention layers and cross-attention layers, from which the prompt-aware visual information with the highest semantic relevance to the analysis target is extracted. The language query will act as one of the input of text encoder, guiding model to output prediction based on the current analysis target.
Considering the pairwise relevance between target span boundaries, we introduce the idea of Energy Based Model~\cite{lecun2006tutorial} to give better span scores and proposed the novel \textit{Energy based Pairwise Expert} that predicts span based on pairing stability.
Experiments on three widely used benchmarks verify that DQPSA outperforms previous approaches and achieves the state-of-the-art performance. A large amount of complementary experiments and ablation studies further demonstrate the effectiveness of components we proposed.

In summary, our contributions are as follows:

\noindent$\bullet$ We proposed \textit{Prompt as Dual Query} module that satisfy diverse focus of different analysis targets on visual information, and strengthen the pairwise relevance between visual information and analysis target. 

\noindent$\bullet$  Inspired by the Energy Based Model (EBM) that quantifying compatibility between variables using an energy scalar, we proposed a novel \textit{Energy based Pairwise Expert} that models the  boundaries pairwise stability of target span.

\noindent$\bullet$  Experiments on three widely used benchmarks Twitter2015, Twitter2017 and Political Twitter shows that DQPSA outperforms previous approaches and achieves SOTA performance. DQPSA also  significantly outperforms Multi-modal Large Language Model (LLM) VisualGLM-6B and Uni-modal LLM ChatGPT-3.5 under fair comparison. 

\section{Related Work}

\subsection{Multi-modal Aspect-Based Sentiment Analysis}

The sentiment analysis task, due to its strong correlation with human sentiment traits, has been regarded as a challenging benchmark to verify the model's semantic comprehension and analytical ability, has attracted many research efforts (~\cite{DBLP:journals/corr/abs-2109-06719}, ~\cite{jing-etal-2021-seeking}).
With the proliferation of multi-modal data disseminated on social media, images are considered to be an important complementary information for sentiment analysis. Thus, MABSA began to receive increasing attention. 

~\cite{DBLP:conf/naacl/WangGJJBWHT22} aligns the image into regional object tags, image-level captions and optical characters as visual contexts for better interactions. 
~\cite{wang-etal-2022-named} injects knowledge-aware information through multi-modal retrieval.
~\cite{cai-etal-2019-multi} treats image attribute features as supplementary modalities to bridge the gap between texts and images. 
~\cite{ju-etal-2021-joint} first proposes
JMASA task that jointly extract aspects and corresponding sentiment polarity to better satisfy the practical applications, 
~\cite{ling-etal-2022-vision} performs various vision-language pre-training tasks to capture crossmodal alignment, and 
~\cite{zhou-etal-2023-aom} designs an aspect-aware attention module to select textual tokens and image blocks that are semantically related to the aspects. 
The above approaches for JMASA either treat MATE and MASC as sequence labelling and binary classification, or require an additional decoding module for sequence generating, and do not emphasize the variability of image focus across analysis targets. Unlike previous works, our proposed DQPSA address MATE and MASC under a unified framework as span recognition, dispensing with the complex sequence generation structure. Meanwhile, we design the \textit{Prompt as Dual Query} module that uses prompt as both visual query and language query, in order to differentially extract prompt-aware visual information and strengthen image-text pairwise relevance.

\subsection{Energy Based Model}
The concept of Energy Based Model was first proposed by~\cite{lecun2006tutorial}. The core idea is to establish a mapping between different variable configurations and a scalar energy that will be able to measure compatibility, thus capturing the dependence between different variable configurations. The target of learning is to find an efficient energy function that maps correct variable configurations to low energy values while mapping incorrect variable configurations to high energy values. The goal of inference is to find variable configurations that minimize the energy. The loss function selected during training can be used to measure the effectiveness of the energy function.
~\cite{zou-etal-2021-unsupervised} introduce the concept of EBM to text-only adversarial domain adaptation to construct an autoencoder during the training phase. This autoencoder maps the source-domain data to a low-energy space and imposes constraints on the target-domain data distribution to align the source-target features. However, due to the fact that the code and dataset used in~\cite{zou-etal-2021-unsupervised} have not been publicly released, and considering the differences in dataset and task, it's hard for us to make a fair comparison with it.

Inspired by the idea of Energy Based Model, we quantify boundary pairing stability of potential span with scalar energies and design the \textit{Energy based Pairwise Expert} to predict span based on pairwise stability. To the best of our knowledge, this is the first attempt that adapting Energy Based Model into MABSA task.

\section{Method}

\begin{figure*}[ht]
\centering 
\includegraphics[width = 0.8\linewidth]{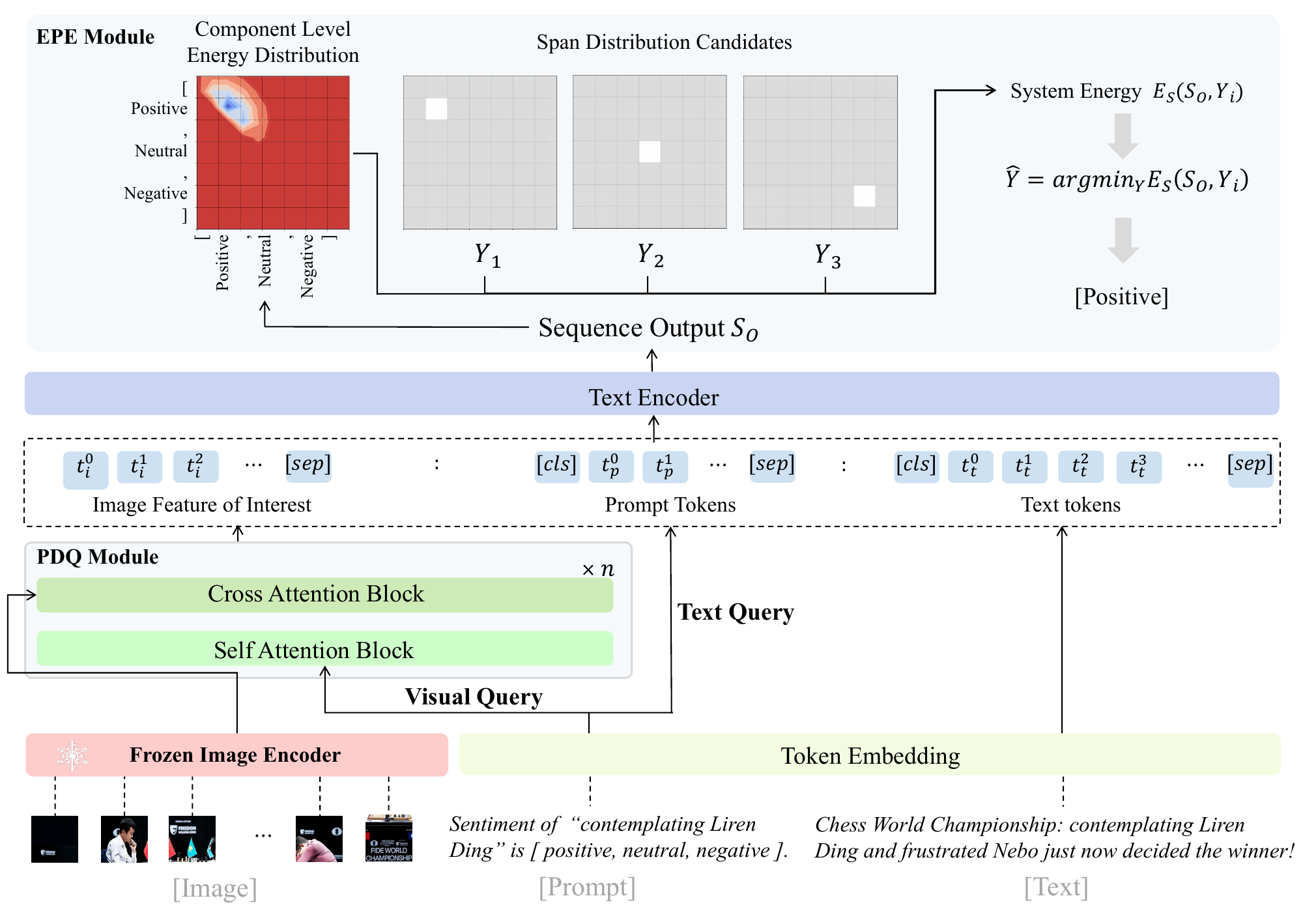}
\caption{The overview of our proposed DQPSA.}
\label{fig:model_overview}
\end{figure*}

In this section, we first introduce the general framework of our proposed DQPSA. Then we introduce the specific structure of \textit{Prompt as Dual Query}, followed by a detailed description of \textit{Energy based Pairwise Expert} .
Figure~\ref{fig:model_overview} demonstrates the framework of our proposed DQPSA, which consists of four main components: a frozen image encoder, a \textit{Prompt as Dual Query} module, a text encoder and an \textit{Energy based Pairwise Expert}. To strengthen the correlation between visual information and analysis target, we design the \textit{Prompt as Dual Query} that allows prompt to interacting with both image and text. Besides, in order to consider the boundary positions of the target span more comprehensively, we design the \textit{Energy based Pairwise Expert} to extract span considering both start and end boundaries simultaneously. In the following subsections, we will illustrate the details of our proposed model.

\subsection{Prompt as Dual Query (PDQ) }

\begin{figure} 
    \centering    
    \includegraphics[width=1\columnwidth]{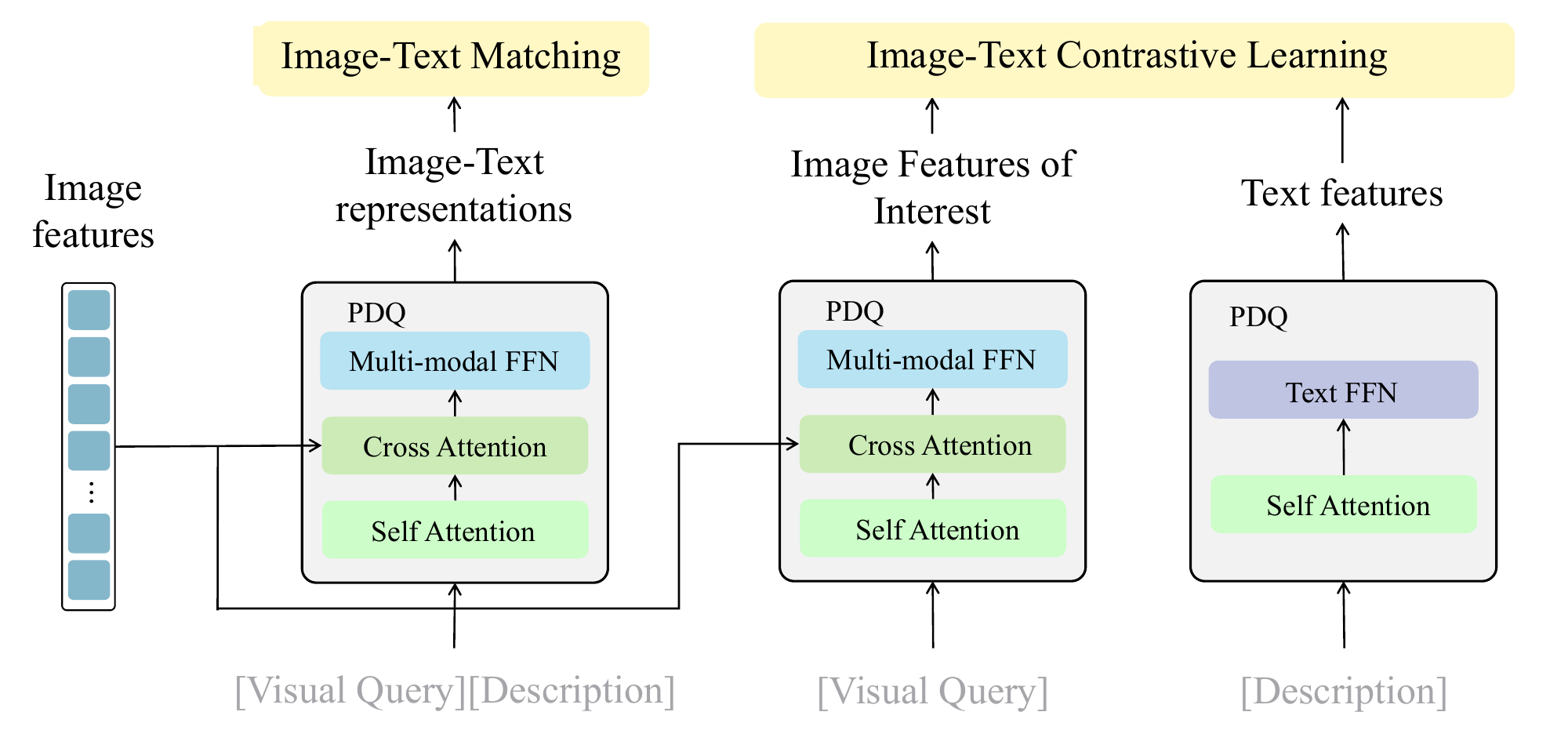}  
    \caption{Demonstration of Prompt as Dual Query}     
    \label{fig:PDQ_stru}     
\end{figure}

We propose the \textit{Prompt as Dual Query} (PDQ) module, which leverage prompt as both visual query and language query, guiding model to focus on different perspectives of visual information and text information according to concrete analysis targets.

Figure~\ref{fig:PDQ_stru} shows the specific structure of PDQ, which is mainly composed of two kinds of blocks stacked alternately: cross-attention block and self-attention block. we assume that the image feature obtained from the frozen image encoder is 
$F_{\textsc{I}}=\left\{I_{\textsc{0}}, I_{\textsc{1}}, I_{\textsc{2}} \cdots I_{L_{\textsc{I}}}\right\}$. 
For each image, we constructed a description for its content, the features of the constructed description and the prompt are 
$F_{\textsc{D}}^{i}=\left\{I_{\textsc{0}}^{i}, I_{\textsc{1}}^{i}, I_{\textsc{2}}^{i} \cdots I_{L_{\textsc{D}}}^{i}\right\}$ 
and $F_{\textsc{P}}^{i}=\left\{I_{\textsc{0}}^{i}, I_{\textsc{1}}^{i}, I_{\textsc{2}}^{i} \cdots I_{L_{\textsc{P}}}^{i}\right\}$,
where $L_{\textsc{I}}$, $L_{\textsc{D}}$, $L_{\textsc{P}}$ are the lengths of the corresponding sequences. $i\in[\textsc{-1},\textsc{N}]$ represents the index of the block in which $F_{\textsc{D}}$ and $F_{\textsc{P}}$ are located and $i=-1$ represents the original word embedding. The PDQ receives three types of text sequences belonging to 
$S=\{F_{\textsc{D}}^{-1},F_{\textsc{P}}^{-1},\mbox{$[F_{\textsc{P}}^{-1}:F_{\textsc{D}}^{-1}]$}\}$ as inputs, with only sequences belonging to 
$\hat{S}=\{F_{\textsc{P}}^{-1},\mbox{$[F_{\textsc{P}}^{-1}:F_{\textsc{D}}^{-1}]$}\}$ 
interacting with image features in cross-attention layer.

The basic formula for attention can be expressed as:
\begin{equation}
    \textsc{Attn}(Q, K, V)=\operatorname{Softmax}\left(\frac{Q \mathrm{~K}^{\top}}{\sqrt{\mathrm{d}_{\mathrm{k}}}}\right) V.
\end{equation}
Suppose the hidden state $H^{i-1}$ of the previous layer, for input $\in S$, the result of self-attention layer can be represented as follows, where $W_{QS},W_{KS},W_{VS},W_{QC},W_{KC},W_{VC}$ are the parameters to be optimized:: 
\begin{equation}
\begin{aligned}
    &\textsc{Self-Attn}(H^{i-1})\\
    &=\textsc{Attn}(W_{\textsc{QS}}H^{i-1}, W_{\textsc{KS}}H^{i-1},W_{\textsc{VS}}H^{i-1}).
\end{aligned}
\end{equation}
As for input $\in \hat{S}$, it will go through an additional cross-attention layer in cross-attention block that can be represented as:
\begin{equation}
\begin{aligned}
    &\textsc{I2T-Attn}(H^{i-1})\\
    &=\textsc{Attn}(W_{\textsc{QC}}H^{i-1}[:L_{\textsc{P}}], W_{\textsc{KC}}F_{\textsc{I}},W_{\textsc{VC}}F_{\textsc{I}}),\\
    &\textsc{Cross-Attn}(H^{i-1})\\
    &=\textsc{cat}[\textsc{I2T-Attn}(H^{i-1}):H^{i-1}[L_{\textsc{P}}:]].
\end{aligned}
\end{equation}
where $H^{i-1}[:L_{\textsc{P}}]$ represents the sub-sequence of $H^{i-1}$ up to the $L_{\textsc{P}}$-th token and $H^{i-1}[L_{\textsc{P}}:]$ represents the sub-sequence of $H^{i-1}$ from and include the $L_{\textsc{P}}$-th token.
In the inference process, we use $F_{\textsc{P}}^{-1}$ as input. Through multiple fusion with $F_{\textsc{I}}$, prompt guides the model to filter the prompt-aware visual information that semantically related to the analysis target.

To further strength the pairwise relevance between visual information and analysis targets, we introduce image-text matching task and in-batch image-text contrastive learning, the specific algorithmic process is as follows:

For image-text matching task, we have
\begin{equation}
\begin{aligned}
    &\textsc{Loss}_{\textsc{ITM}}=-\frac{1}{2}\sum_{i=0}^{1}p_{i}log(q_{i}), \\
    &p=\textsc{mean}((W_{\textsc{ITM}}[F_{\textsc{VQ}}^{\textsc{N}}:F_{\textsc{D}}^{\textsc{N}}])[:L_{\textsc{VQ}}]).
\end{aligned}
\end{equation}
where the hidden state in the last block of PDQ will go though an linear projection and we select the average of tokens corresponding to visual query as model prediction $p$.
$q$ is the label that identifies whether the image and description match or not. $W_{\textsc{ITM}}$ is the parameter to be optimized.

For in-batch image-text comparative learning, we firstly use the first token of $F_{\textsc{VQ}}^{\textsc{N}}$ and $F_{\textsc{D}}^{\textsc{N}}$ as $\textsc{[CLS]}$ token to construct the similarity matrix between different $F_{\textsc{VQ}}^{\textsc{N}}$ and $F_{\textsc{D}}^{\textsc{N}}$ within the same batch:
\begin{equation}
\begin{aligned}
   I^{\textsc{ITC}}&=[I^{\textsc{ITC}}_{\textsc{1}}, I^{\textsc{ITC}}_{\textsc{2}} \cdots I^{\textsc{ITC}}_{\textsc{B}}],\\
   T^{\textsc{ITC}}&=[T^{\textsc{ITC}}_{\textsc{1}}, T^{\textsc{ITC}}_{\textsc{2}} \cdots T^{\textsc{ITC}}_{\textsc{B}}].\\
\end{aligned}
\end{equation}
where $\textsc{B}$ is the batch size, $I^{\textsc{ITC}}_{i}$ and $T^{\textsc{ITC}}_{i}$ is the $\textsc{[CLS]}$ token of the $i$-th $F_{\textsc{VQ}}^{\textsc{N}}$ and $F_{\textsc{D}}^{\textsc{N}}$ in batch.
And then we construct the similarity vector $p_{\textrm{i2d}}\ p_{\textrm{d2i}}$,  and corresponding label $q_{\textrm{i2d}}\ q_{\textrm{d2i}}$, for the $i$-th $F_{\textsc{VQ}}^{\textsc{N}}$ and $F_{\textsc{D}}^{\textsc{N}}$ in the batch:
\begin{equation}
\begin{aligned}
   &p^{\textrm{i2d}}=T^{\textsc{ITC}^{\top}}I^{\textsc{ITC}}_{i},p^{\textrm{d2i}}=I^{\textsc{ITC}^{\top}}T^{\textsc{ITC}}_{i},\\
   &q^{\textrm{i2d}},q^{\textrm{d2i}}\in R^{\textsc{B}\times \textsc{1}},q^{\textrm{i2d}}_{j},q^{d2i}_{j}=1\ if\ j==i\ else\ 0.
\end{aligned}
\end{equation}
finally, the $\textsc{Loss}_{\textsc{ITC}}$ can be represented as:
\begin{equation}
\begin{aligned}
   &\textsc{Loss}_{\textsc{ITC}}=-\frac{1}{\textsc{B}} (\sum_{j=1}^{\textsc{B}}p^{\textrm{i2d}}_{j}log(q^{\textrm{i2d}}_{i})+\sum_{j=1}^{\textsc{B}}p^{\textrm{d2i}}_{j}log(q^{\textrm{d2i}}_{i})).
\end{aligned}
\end{equation}
Optimizing $\textsc{Loss}_{\textsc{ITM}}$ and $\textsc{Loss}_{\textsc{ITC}}$ enables model to capture the pairwise relevance of image-text pairs thus obtain the capability of prompt-aware visual information extraction.

\subsection{Energy based Pairwise Expert (EPE)}

\begin{figure}[t] 
    \centering    
    \includegraphics[width=1\columnwidth]{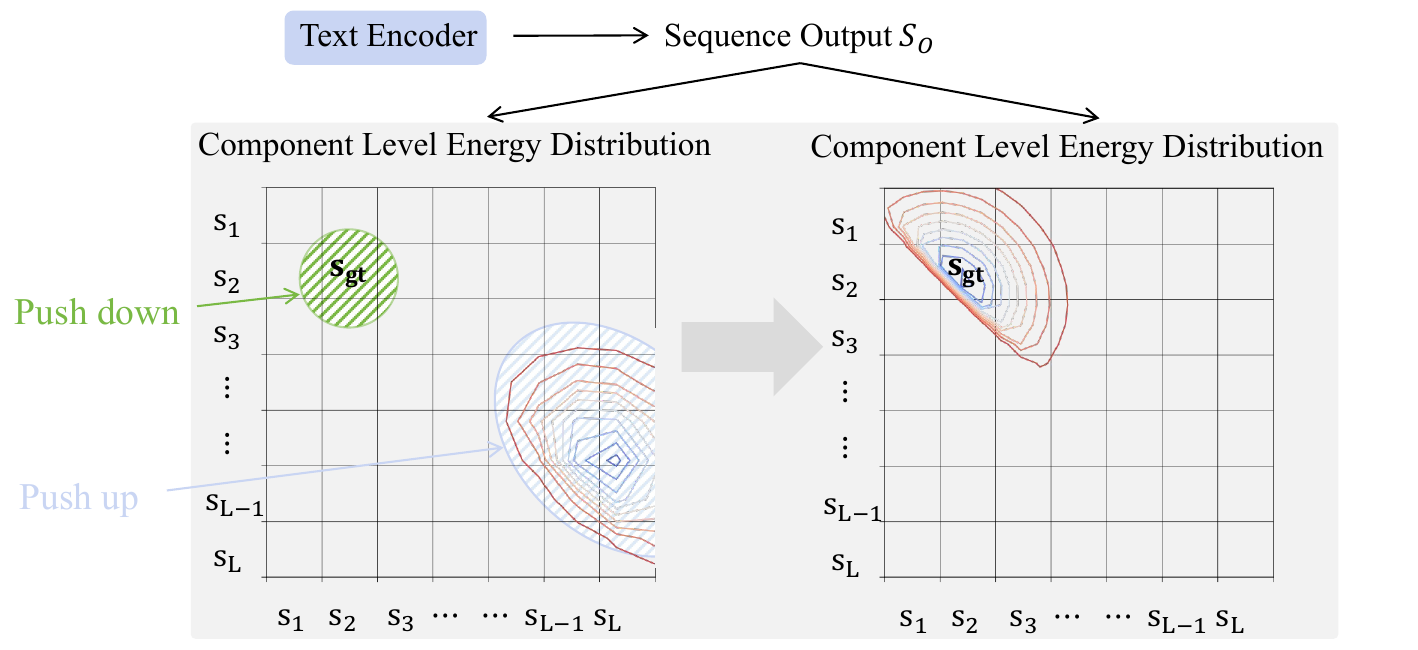}  
    \caption{Demonstration of Energy based Pairwise Expert}     
    \label{fig:EPE_stru}     
\end{figure}


To capture the semantical relevance of start and end boundary within a span, we introduce the idea of Energy Based Model~\cite{lecun2006tutorial} to give better span scores and proposed the novel \textit{Energy based Pairwise Expert} that predicts span based on pairwise stability.

EBM captures the correlation between variables by creating a mapping of variable combinations to a scalar energy. In the inference phase, the model aims to find the combination of variables that minimizes the energy of the system, in the training phase, the model is forced to find an energy function that assigns lower energy values to paired combinations of variables while assigning higher energy values to unpaired combinations of variables.

The core theory of EBM is learning an energy function that assigns lower energy values to paired combinations of variables while assigning higher energy values to unpaired combinations of variables.

Following the theory of EBM, we try to map the boundary pairing relations of a target span to a scalar energy, which in turn quantitatively describe the intensity of the pairing relations at the boundary of a given span. 
Figure~\ref{fig:EPE_stru} demonstrates the learning process of EPE, in which we lower the energy of positions with pairwise relevance while raising the energy of others.
Specifically, we design the \textit{Component-level energy function} $E_{\textsc{C}}$ and \textit{System-level energy function} $E_{\textsc{S}}$. 
Let the output sequence of the text encoder be $S_{\textsc{O}}=\{s_{\textsc{1}}, s_{\textsc{2}} \cdots s_{\textsc{L}}\}$, $\textsc{L}$ is the length of $S_{\textsc{O}}$. We use $E_{\textsc{C}}$ to denote the pairing energy of the span whose boundaries are the $i$-th and $j$-th token, denoted as 
\begin{equation}
\begin{aligned}
   E_{\textsc{C}}(W_{\textsc{S}},W_{\textsc{E}},s_{i},s_{j})=-(s_{i}^{\top}W_{\textsc{S}_{i}}^{\top}(W_{\textsc{E}_{j}}s_{j})).
\end{aligned}
\end{equation}
where $W_{\textsc{S}}$ and $W_{\textsc{E}}$ are the parameters to be optimized. The smaller $E_{\textsc{C}}$ represents the more stable pairing relation, i.e., the higher the probability that $S_{\textsc{O}}[i:j+1]$ is the target span.

For the distribution of predicted span over the entire sequence, we use $E_{\textsc{S}}$ to measure the energy of the entire system. Let the span distribution matrix $Y\in R^{\textsc{L}\times \textsc{L}}$ be the label of the target span, where $y_{ij}\in \{0,1\}$ indicates whether $S_{\textsc{O}}[i:j+1]$ is the target span or not. The system level energy can be expressed as
\begin{equation}
\begin{aligned}
   E_{\textsc{S}}(S_{\textsc{O}},Y)=\sum_{i=0}^{\textsc{L}}\sum_{j=i}^{\textsc{L}}(E_{\textsc{C}}(s_{i},s_{j})y_{ij}).
\end{aligned}
\end{equation}

In the inference phase, we select the span distribution matrix that minimizes the energy of the system. In the training phase, we construct the loss function directly using $E_{\textsc{C}}$ as 
\begin{equation}
\begin{aligned}
   \textsc{Loss}_{\textsc{EPE}}&=-\frac{1}{\frac{\textsc{L}(\textsc{L}+1)}{2}}\sum_{i=0}^{\textsc{L}}\sum_{j=i}^{\textsc{L}}\bigg (y_{ij}log(x_{ij})\\
   &+(1-y_{ij})log(1-x_{ij})\bigg ),\\
   &x_{ij}=\operatorname{Sigmoid}(-E_{\textsc{C}}(s_{i},s_{j})).
\end{aligned}
\end{equation}
by optimizing $\textsc{Loss}_{\textsc{EPE}}$, the model will learn the energy function $E_{\textsc{C}}$ that allocates lower energy to the paired systems $(S_{\textsc{O}},Y)$ while allocating higher energy to the unpaired ones.

Based on all the above narratives, the final training loss of the model is represented as
\begin{equation}
\begin{aligned}
\textsc{Loss}=\lambda_{1}\textsc{Loss}_{\textsc{ITM}}+\lambda_{2}\textsc{Loss}_{\textsc{ITC}}+\lambda_{3}\textsc{Loss}_{\textsc{EPE}}.
\end{aligned}
\end{equation}
where $\lambda_{1},\lambda_{2},\lambda_{3}$ are predefined hyper-parameters

\begin{table*}
\centering
\resizebox{0.9\linewidth}{!}{
\begin{tabular}{clccccccc}
\toprule
 & \multirow{2}{*}{\bf Methods}  & \multicolumn{3}{c}{\bf Twitter2015} & & \multicolumn{3}{c}{\bf Twitter2017}  \\
\cmidrule{3-5} \cmidrule{7-9} 
&  & P & R & F1 & & P & R & F1 \\
\midrule
\multirow{3}{*}{Text-based} & SPAN~\cite{hu-etal-2019-open} &53.7 &53.9 &53.8 & & 59.6 &61.7 &60.6 \\
& D-GCN~\cite{chen-etal-2020-joint-aspect} & 58.3& 58.8 &59.4 & &64.2 &64.1 & 64.1 \\
& BART~\cite{yan-etal-2021-unified} & 62.9 & 65.0 & 63.9 & & 65.2 & 65.6 & 65.4 \\
\midrule
\multirow{9}{*}{Multi-modal} & UMT+TomBERT~\cite{yu-etal-2020-improving-multimodal,DBLP:conf/ijcai/Yu019} & 58.4 &61.3 & 59.8 & & 62.3 & 62.4 &62.4 \\
& OSCGA+TomBERT~\cite{DBLP:conf/mm/WuZCCL020,DBLP:conf/ijcai/Yu019} & 61.7 & 63.4 & 62.5 & & 63.4 & 64.0 & 63.7 \\
& OSCGA-collapse~\cite{DBLP:conf/mm/WuZCCL020} & 63.1 &63.7&63.2 &&63.5&63.5&63.5\\
& RpBERT-collapse~\cite{DBLP:conf/aaai/0006W0SW21} & 49.3 &46.9 &48.0& &57.0&55.4&56.2\\
& UMT-collapse~\cite{yu-etal-2020-improving-multimodal} &61.0 &60.4 & 61.6 & & 60.8 & 60.0 &61.7 \\
& JML~\cite{ju-etal-2021-joint}&65.0&63.2&64.1 & &66.5 & 65.5 & 66.0\\
& VLP-MABSA~\cite{ling-etal-2022-vision} &65.1 &68.3 & 66.6 & & 66.9 & 69.2 & 68.0\\
& CMMT~\cite{DBLP:journals/ipm/YangNY22} & 64.6 &68.7 & 66.5 &  &67.6 & 69.4 & 68.5 \\
& AoM~\cite{zhou-etal-2023-aom} & 67.9&  69.3 & 68.6 &&  68.4 & \textbf{71.0} & 69.7 \\
& \bf DQPSA (ours) & \textbf{71.7}&  \textbf{72.0} & \textbf{71.9} &&  \textbf{71.1} & 70.2 & \textbf{70.6} \\
\bottomrule

\end{tabular}
}
\caption{Results of Twitter2015 and Twitter2017, JMASA task. The best results are bold-typed.}
\label{tab:JMASA_15_17_res}
\end{table*}

\begin{table}[t]
\centering
\resizebox{0.9\linewidth}{!}{
\begin{tabular}{lccc}
\toprule
 \multirow{2}{*}{\bf Methods} & \multicolumn{3}{c}{Political-Twitter} \\
 \cmidrule{2-4}
&  P & R & F1 \\
\midrule
RoBERTa~\cite{DBLP:journals/corr/abs-1907-11692} &63.1&62.1&62.6 \\
UMT+collapse~\cite{yu-etal-2020-improving-multimodal} &54.9&54.7&54.8 \\
JML~\cite{ju-etal-2021-joint} &63.6&59.4&61.4\\
UMT-RoBERTa~\cite{yu-etal-2020-improving-multimodal,DBLP:journals/corr/abs-1907-11692} &63.8&63.4&63.6   \\
JML-PoBERTa~\cite{ju-etal-2021-joint,DBLP:journals/corr/abs-1907-11692} &63.0&60.2&61.6\\
CMMT~\cite{DBLP:journals/ipm/YangNY22}  &65.3&\textbf{65.7}&65.5\\
\bf DQPSA (ours) & \textbf{68.3} & 65.5 & \textbf{66.9}   \\ 
\bottomrule
\end{tabular}
}
\caption{Results of Political Twitter, JMASA task. The best results are bold-typed. }
\label{tab:JMASA_political_res}

\end{table}

\section{Experiments}

\subsection{Experimental settings}

\subsubsection{Dataset}
Following previous works, we adopt two widely used benchmarks: Twitter2015 and Twitter2017~\cite{DBLP:conf/ijcai/Yu019} to evaluate our proposed DQPSA. Besides, we employ another Political Twitter dataset\footnote{Political Twitter for evaluaton is consistence with dataset released by~\cite{DBLP:journals/ipm/YangNY22}} from~\cite{DBLP:conf/iconference/YangYZN21} for JMASA task. In pre-training stage, we use COCO2017 dataset and ImageNet dataset.

\subsubsection{Implementation Details}
In our proposed model, we choose CLIP-ViT-bigG-14-laion2B-39B-b160k~\cite{DBLP:conf/icml/RadfordKHRGASAM21, ilharco_gabriel_2021_5143773} as the frozen image encoder, FSUIE-base~\cite{peng-etal-2023-fsuie} as the text encoder, \textit{Prompt as Dual Query} module is based on BERT-base~\cite{devlin-etal-2019-bert} architecture and pre-training parameter and randomly initialized cross-attention layers. Refer to Appendix A for detailed information of our backbone and selections of hyper-parameters.

We first employ a two-stage pre-training with $5$ epochs for each stage. Then We trained model for $50$ epochs with an AdamW optimizer on the datasets of each task, and selected the final model based on the performance on the development set. Associated code and pre-trained models will be made publicly available upon receipt.

\subsubsection{Evaluation Metrics}
Following previous work, we evaluate the performance of our model on the MATE and JMASA tasks with Precision (P), Recall (R) and Micro-F1 (F1) score, while on MASC task, we report Accuracy (Acc) and Micro-F1 (F1) score for comparison.

\subsubsection{Pre-training}
In order to equip the PDQ with initial capability of prompt-controlled image comprehending, we employed a two-stage pre-training before adapting to the specific MABSA task. 
For ImageNet data, we train model to predict the entity class contained in the image under the guidance of prompt, which helps the model to capture the word level pairwise relevance between images and entities. 
For COCO data, the model is trained to predict the descriptions that are relevant to the content of image under the guidance of prompt, from which the model will learn the sentence level pairwise relevance between image and text.
See Appendix B for specific constructs of prompt, description and text.

During phase 1, to prevent the initialized PDQ from detracting the semantic comprehension of text encoder, we freeze all parameters except PDQ and EPE. While for subsequent training, we train all model parameters except image encoder.

\subsection{Main Results}
\begin{table}[t]
\centering
\resizebox{1\linewidth}{!}{
\begin{tabular}{lccccccc}
\toprule
\multirow{2}{*}{\bf Methods} & \multicolumn{3}{c}{Twitter2015} & & \multicolumn{3}{c}{Twitter2017} \\
 \cmidrule{2-4} \cmidrule{6-8}
&  P & R & F1 & & P & R & F1\\
\midrule
RAN~\cite{DBLP:conf/nlpcc/WuCWLC20} &80.5 &81.5 &81.0 && 90.7 &90.7&90.0\\
UMT~\cite{yu-etal-2020-improving-multimodal} &77.8 & 81.7 & 79.7&& 86.7 &86.8&86.7\\
OSCGA~\cite{DBLP:conf/mm/WuZCCL020} & 81.7 & 82.1 & 81.9 && 90.2& 90.7 & 90.4\\
JML~\cite{ju-etal-2021-joint} &83.6 &81.2 & 82.4& &92.0 & 90.7 & 91.4 \\
VLP-MABSA~\cite{ling-etal-2022-vision} & 83.6 & 87.9 &85.7& &90.8 & 92.6  & 91.7  \\
CMMT~\cite{DBLP:journals/ipm/YangNY22} & 83.9 &\textbf{88.1} &85.9& & 92.2 &\textbf{93.9} &93.1 \\
AoM~\cite{zhou-etal-2023-aom}  & 84.6 & 87.9 & 86.2  & &  91.8  &  92.8 &  92.3 \\ 
\bf DQPSA (ours) & \textbf{88.3} & 87.1 & \textbf{87.7}   & &  \textbf{95.1}  &  93.5 &  \textbf{94.3} \\ 
\bottomrule
\end{tabular}
}
\caption{Results of Twitter2015 and Twitter2017, MATE task. The best results are bold-typed.}
\label{tab:MATE_15_17_res}
\end{table}

\subsubsection{Results of JMASA Task}
Table~\ref{tab:JMASA_15_17_res} and ~\ref{tab:JMASA_political_res} show the results of JMASA task. It can be seen that, by introducing the \textit{Prompt as Dual Query} module and the \textit{Energy based Pairwise Expert} module, DQPSA significantly outperforms the sub-optimal models (3.3 on Twitter2015 and 0.9 on Twitter2017) and achieves SOTA results. This demonstrates the effectiveness of differentially leveraging visual information according to different analytical targets and focusing on the pairwise relevance of the target span.

Compared to the text-based models, the multi-modal models perform better in general and DQPSA far exceeds all text-based models. This verifies that image modal introduced by our method does provide important supplementary information for sentiment analysis. 

Compare to the span based methods, EPE improves the span boundary recognition process. Instead of separately predicting the start and end boundaries of span, EPE capture the pairwise relevance of span boundaries, which provides a better understanding of the distribution of target spans. 
Unlike directly using visual features as a prefix to textual inputs, PDQ module filters out visual noise and selects visual information beneficial to the current task. It also bridge the modality gap between visual and textual features through stacked attention operation.
Furthermore, our method focuses on task-specific visual information rather than token-specific visual information, allowing for a more macro-level utilization of visual information.
Compare to the span based methods, EPE improves the span boundary recognition process. Instead of separately predicting the start and end boundaries of span, EPE capture the pairwise relevance of span boundaries, which provides a better understanding of the distribution of target spans. 
Unlike directly using visual features as a prefix to textual inputs, PDQ module filters out visual noise and selects visual information beneficial to the current task. It also bridge the modality gap between visual and textual features through stacked attention operation.
Furthermore, our method focuses on task-specific visual information rather than token-specific visual information, allowing for a more macro-level utilization of visual information.

In contrast to approaches that use collapse labels or work with pipelines using different models, our approach uses a unified framework for the MASC and MATE tasks, completely eliminating the formal differences between two tasks, helping the model to learn the interactions between the two sub-tasks in terms of semantic information and sentiment with a more concise structure, resulting in better performance. 
Compared with methods that focus on token-various visual information, our approach focuses on target-various visual information to filter and leverage image features from a more macroscopic perspective, achieving better performance while reducing computational effort.

\begin{table}[h]
\centering
\resizebox{0.9\linewidth}{!}{
\begin{tabular}{lccccc}
\toprule
\multirow{2}{*}{\bf Methods}& \multicolumn{2}{c}{Twitter2015} & & \multicolumn{2}{c}{Twitter2017} \\
\cmidrule{2-3} \cmidrule{5-6}
&  ACC & F1 & &  ACC & F1\\
\midrule
ESAFN~\cite{DBLP:journals/taslp/YuJX20} & 73.4 & 67.4 & &67.8 & 64.2\\
TomBERT~\cite{DBLP:conf/ijcai/Yu019}& 77.2 &71.8 & & 70.5 & 68.0 \\
CapTrBERT~\cite{DBLP:conf/mm/0001F21} & 78.0 & 73.2 &  &72.3 & 70.2 \\
JML~\cite{ju-etal-2021-joint} & 78.7 & - & & 72.7 & - \\
VLP-MABSA~\cite{ling-etal-2022-vision} & 78.6 & 73.8 & &73.8 &71.8 \\
CMMT~\cite{DBLP:journals/ipm/YangNY22} & 77.9 & - & & 73.8 & - \\
AoM~\cite{zhou-etal-2023-aom}  & 80.2 & 75.9 & &\textbf{76.4} & \textbf{75.0} \\
\bf DQPSA (ours) & \textbf{81.1} & \textbf{81.1} & & 75.0 & \textbf{75.0} \\

\bottomrule
\end{tabular}
}
\caption{Results of different methods for MASC.}
\label{tab:MASC_15_17_res}
\end{table}

\subsubsection{Results of MATE and MASC Task}
Table~\ref {tab:MATE_15_17_res} and~\ref {tab:MASC_15_17_res} show the results of the MATE and MASC task. Consistent with the results of JMASA task, DQPSA also achieves a significant improvement or competitive performance in the two sub-tasks performance.

Specifically, DQPSA outperforms the sub-optimal method in the MATE task by 1.5 on Twitter2015 and 2.0 on Twitter2017, which suggests that our method helps model to focus on the image information related to the aspects and filter out irrelevant information.

As for MASC, we note that the model made a huge improvement on Twitter2015 (5.2 over previous SOTA), but only achieved competitive results on Twitter2017. On the one hand, we think it is because the sentiment analysis data in Twitter2017 is more challenging
for it contains a significant number of unresolvable and unidentifiable symbols, including emojis commonly used on Twitter, which posed a relative hard challenge for DQPSA,
on the other hand, model with large scale may better capture the correlation between aspect and sentiment in difficult cases. However, considering the difference in the number of trainable parameters, the results of MASC are still convincing in proving the effectiveness of our approach.

\subsection{Ablation Study}

\begin{table}[t]
\centering
\resizebox{1\linewidth}{!}{
\begin{tabular}{lccccccc}
\toprule
\multirow{2}{*}{\bf Methods} & \multicolumn{3}{c}{Twitter2015} & & \multicolumn{3}{c}{Twitter2017} \\
 \cmidrule{2-4} \cmidrule{6-8}
&  MATE & MASC & JMASA & & MATE & MASC & JMASA\\
\midrule
DQPSA           &87.7  &81.1  &71.9  &&94.3 &75.0 &70.6\\
w/o EPE         &86.3  &80.9  &69.1  &&92.5 &73.1 &66.8\\
w/o PDQ         &87.4  &78.4  &69.9  &&93.8 &69.7 &65.6\\
w/o PDQ\&EPE    &84.4  &76.4  &63.3  &&90.8 &68.6 &64.1\\
\bottomrule
\end{tabular}
}
\caption{Results of Ablation Study}
\label{tab:ablation_study_res}
\end{table}

To further investigate the contribution of each component to model performance improvement, we conducted ablation studies on Twitter2015 and Twitter2017 for MATE, MASC, and JMASA tasks, and the results of F1 scores are illustrated in table~\ref{tab:ablation_study_res}. To examine the effect of \textit{Energy based Pairwise Expert} module, we follow the original setting of text encoder to predict positions of start and end boundaries independently. To examine the effect of \textit{Prompt as Dual Query} module, we replace the visual query with a set of optimizable tokens.

It can be seen that removing both the PDQ and the EPE detracts somewhat from the model's performance. Specifically, the model without PDQ achieves competitive performance on the MATE task, but performance on the MASC and JMASA tasks drops significantly. This is because the prompt we constructed for the MASC task is different among potential aspects, and optimizable tokens are not sufficient to capture this target-variability.
However, the prompt used by MATE is relatively fixed, so optimizable tokens can fit the target requirements of MATE to a certain extent and acts as soft prompt. The JMASA task, as a prolongation of the two subtasks, naturally shows a decrease in performance due to decrease on MATE and MASC. And this verifies that our \textit{Prompt as Dual Query} can satisfy the differential focus on visual information of different targets in the MABSA topic.

For model without EPE, we can see that the model shows a significant performance degradation on all three tasks. This is due to the fact that EPE, as a module for the model to make span decisions, has an auxiliary effect on all the tasks performed by the model. By introducing EPE, the model does not consider the boundaries of the target span in isolation, but makes full use of the pairwise relevance between the boundaries of the spans, which further enriches the knowledge learnt by the model, and thus improves the effect on multiple tasks.

Considering the whole table, the introduction of PDQ and EPE alone can significantly improve the model performance, in which EPE leads to comprehensive performance improvement by capture the span pairwise relevance, while PDQ focuses more on guiding the model to filter visual information according to different target requirements thus is more effective in enhancing fine-grained MASC tasks. Simultaneous application of the two will further boost the model performance.

\subsection{Case Study}
For in-depth analysis, we also trained a PSA model based on FSUIE-base without using \textit{Prompt as Dual Query} module and only receiving text as input. We apply PSA model to different case study as follows:

\subsubsection{EPE on Sentiment Analysis of Plain Text}

To verify the robustness of \textit{Energy based Pairwise Expert}, we apply PSA to the ASTE-Data-V2~\cite{xu-etal-2020-position} of Aspect Sentiment Triplet Extraction (ASTE) task, and compare it with the best result of the existing work FSUIE-base~\cite{peng-etal-2023-fsuie}. Table~\ref{tab:ASTE_res} shows the F1 scores of PSA on the ASTE task, it shows that EPE delivers a huge performance improvement on all the four datasets and achieves the up-to-date SOTA results. This demonstrates the extensibility and robustness of our proposed EPE, as well as its strong performance under span recognition especially in cases involving sentiment analysis.

\begin{table}[t]
\small
\centering
\begin{tabular}{lcccc}
\toprule
\bf Methods&  14lap & 14res &15res &16res \\
\midrule
FSUIE-base &65.6 &74.1 &70.6 &75.8\\
\bf PSA (ours) &69.8 &78.5 &78.3 &79.2\\

\bottomrule
\end{tabular}
\caption{Results on ASTE-DATA-V2 datasets (14lap, 14res, 15res, and 16res)}
\label{tab:ASTE_res}
\end{table}

\subsubsection{W/o Information from Image Modal}

\begin{table}
\centering
\small
\resizebox{1\linewidth}{!}{
\begin{tabular}{lccccccc}
\toprule
 \multirow{2}{*}{\bf Methods}& \multicolumn{3}{c}{Twitter2015} & & \multicolumn{3}{c}{Twitter2017} \\
 \cmidrule{2-4} \cmidrule{6-8}
&  MATE & MASC & JMASA & & MATE & MASC & JMASA\\
\midrule
DQPSA  &87.7  &81.1  &71.9  &&94.3 &75.0 &70.6\\
PSA    &80.8  &77.7  &62.6  &&91.2 &68.6 &61.6\\
\bottomrule
\end{tabular}
}
\caption{Results of models w \& w/o image modal}
\label{tab:w_w/o_image_res}
\end{table}

To investigate whether our approach helps the model utilize visual information more efficiently, we apply PSA comparing to DQPSA on Twitter2015 and Twitter2017 to examine the effects of the introduction of image modal on the model performance. Table~\ref{tab:w_w/o_image_res} reports F1 scores for PSA and DQPSA on Twitter2015 and Twitter2017 for the three tasks. 

It can be seen from table~\ref{tab:w_w/o_image_res} that: with \textit{Prompt as Dual Query} module 
, our DQPSA significantly outperformed PSA that without image modal on a variety of tasks. This verifies that our proposed method is efficient in helping models to exploit the rich information from images.

\subsubsection{Sentiment analysis compared to LLMs}
\begin{table}[t]
\centering
\small
\resizebox{1\linewidth}{!}{
\begin{tabular}{lccccccc}
\toprule
 \multirow{2}{*}{\bf Models}& \multicolumn{3}{c}{Twitter2015} & & \multicolumn{3}{c}{Twitter2017} \\
 \cmidrule{2-4} \cmidrule{6-8}
&  P & R & F1 & & P & R & F1\\
\midrule
\bf DQPSA (ours)  &81.1  &81.1  &81.1  &&75.0 &75.0 &75.0\\
VisualGLM-6B   &69.2  &64.6 &66.8 &&57.2 &52.0 &54.5\\
ChatGPT-3.5 &66.3 &66.3 &66.3 & &58.9 &58.9 &58.9\\
\bottomrule
\end{tabular}
}
\caption{Results of comparison with LLMs on MASC task}
\label{tab:LLM_Sentiment analysis_results}
\end{table}

Recently, Large Language Models (~\cite{DBLP:journals/corr/abs-2307-00360}, ~\cite{du2022glm}, etc) have shown impressive performance in a wide range of NLP tasks with their powerful language perception and generation capabilities.
In order to verify whether our model has an advantage over large models on the MABSA task, we compare the model's performance with that of common LLMs include VisualGLM-6B and ChatGPT-3.5. Since LLMs are not designed for identifying aspects in text and it is more difficult to unify the output structure, we only test them on the MASC task for a fair comparison.
Table~\ref{tab:LLM_Sentiment analysis_results} shows the results of DQPSA and other LLMs on the MASC task. The results show that our DQOSA using fewer parameters obtains significantly better performance than LLMs, which also verifies the superiority and effectiveness of our proposed DQPSA framework.
It should be noted that due to the lack of a visual module, ChatGPT 3.5 only receives text as input. Therefore, despite its impressive generation capability, the performance of ChatGPT3.5 is not very outstanding. This also verifies that there is a great potential and importance for the integration and utilization of information from different modalities in multi-modal tasks.

\subsection{Visualization}
\begin{figure}[t] 
    \centering    
    \includegraphics[width=1.05\columnwidth]{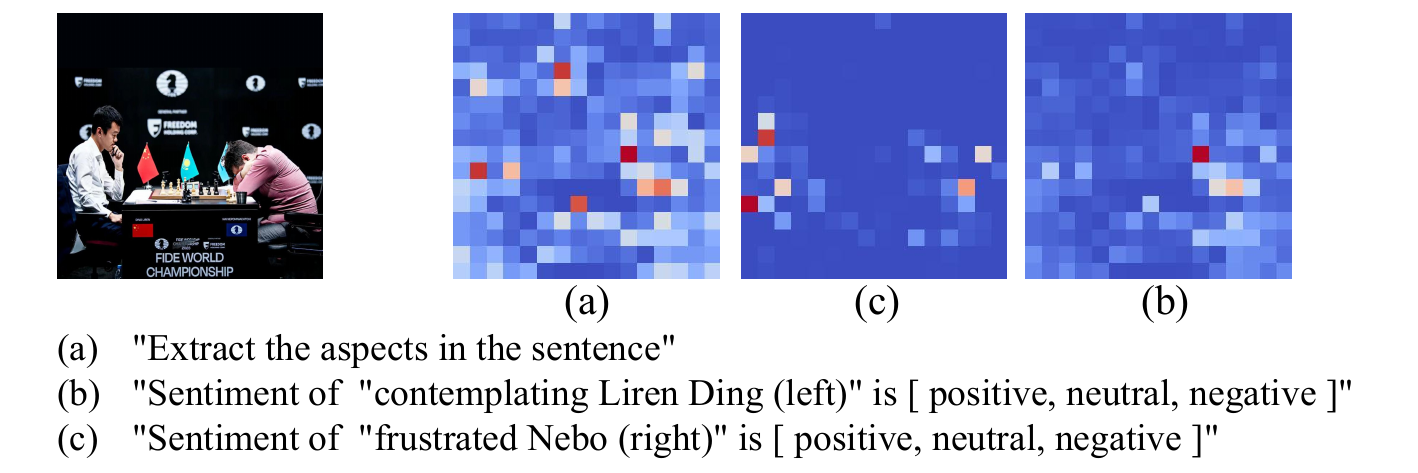}  
    \caption{Visualization of Prompt as Dual Query}     
    \label{fig:attn_vis}     
\end{figure}

To further validate the effectiveness of our proposed method, we visualize the attention matrices in the last cross-attention layer to verify if \textit{Prompt as Dual Query} module guide model to focus on different visual information based on various analysis targets. Figure~\ref{fig:attn_vis} presents the visualization results along with the corresponding prompts. 
It can be observed that for the MATE task, model exhibits a relatively uniform distribution of attention over the image, indicating that model analyzes potential aspects by incorporating a larger receptive field. However, when analyzing different aspects of sentiment, model demonstrates distinct focus towards different regions of visual information. This result once again confirms that our proposed PDQ effectively captures the variations in the focus of visual information across different analysis targets.

\section{Conclusion}
In this paper, we proposed a novel framework, named DQPSA, for Multi-modal Aspect-Based Sentiment Analysis (MABSA). 
We use a well-designed \textit{Prompt as Dual Query} module that leveraging prompt as both visual query and language query to extract the prompt-aware visual information, thus satisfying various focus of different analysis targets on the visual information. 
Besides, we capture the boundaries pairing of analysis target with the perspective of Energy based Model and predict span based on pairwise stability with an \textit{Energy base Pairwise Expert} module. 
Performance on three widely used benchmark datasets verifies that our method outperforms previous methods.


\bibliography{aaai23}

\begin{thebibliography}{34}
\providecommand{\natexlab}[1]{#1}

\bibitem[{Cai, Cai, and Wan(2019)}]{cai-etal-2019-multi}
Cai, Y.; Cai, H.; and Wan, X. 2019.
\newblock Multi-Modal Sarcasm Detection in {T}witter with Hierarchical Fusion Model.
\newblock In \emph{Proceedings of the 57th Annual Meeting of the Association for Computational Linguistics}, 2506--2515. Florence, Italy: Association for Computational Linguistics.

\bibitem[{Chen, Tian, and Song(2020)}]{chen-etal-2020-joint-aspect}
Chen, G.; Tian, Y.; and Song, Y. 2020.
\newblock Joint Aspect Extraction and Sentiment Analysis with Directional Graph Convolutional Networks.
\newblock In \emph{Proceedings of the 28th International Conference on Computational Linguistics}, 272--279. Barcelona, Spain (Online): International Committee on Computational Linguistics.

\bibitem[{Devlin et~al.(2019)Devlin, Chang, Lee, and Toutanova}]{devlin-etal-2019-bert}
Devlin, J.; Chang, M.-W.; Lee, K.; and Toutanova, K. 2019.
\newblock {BERT}: Pre-training of Deep Bidirectional Transformers for Language Understanding.
\newblock In \emph{Proceedings of the 2019 Conference of the North {A}merican Chapter of the Association for Computational Linguistics: Human Language Technologies, Volume 1 (Long and Short Papers)}, 4171--4186. Minneapolis, Minnesota: Association for Computational Linguistics.

\bibitem[{Du et~al.(2022)Du, Qian, Liu, Ding, Qiu, Yang, and Tang}]{du2022glm}
Du, Z.; Qian, Y.; Liu, X.; Ding, M.; Qiu, J.; Yang, Z.; and Tang, J. 2022.
\newblock GLM: General Language Model Pretraining with Autoregressive Blank Infilling.
\newblock In \emph{Proceedings of the 60th Annual Meeting of the Association for Computational Linguistics (Volume 1: Long Papers)}, 320--335.

\bibitem[{Hu et~al.(2019)Hu, Peng, Huang, Li, and Lv}]{hu-etal-2019-open}
Hu, M.; Peng, Y.; Huang, Z.; Li, D.; and Lv, Y. 2019.
\newblock Open-Domain Targeted Sentiment Analysis via Span-Based Extraction and Classification.
\newblock In \emph{Proceedings of the 57th Annual Meeting of the Association for Computational Linguistics}, 537--546. Florence, Italy: Association for Computational Linguistics.

\bibitem[{Ilharco et~al.(2021)Ilharco, Wortsman, Wightman, Gordon, Carlini, Taori, Dave, Shankar, Namkoong, Miller, Hajishirzi, Farhadi, and Schmidt}]{ilharco_gabriel_2021_5143773}
Ilharco, G.; Wortsman, M.; Wightman, R.; Gordon, C.; Carlini, N.; Taori, R.; Dave, A.; Shankar, V.; Namkoong, H.; Miller, J.; Hajishirzi, H.; Farhadi, A.; and Schmidt, L. 2021.
\newblock OpenCLIP.
\newblock If you use this software, please cite it as below.

\bibitem[{Jing et~al.(2021)Jing, Li, Zhao, and Jiang}]{jing-etal-2021-seeking}
Jing, H.; Li, Z.; Zhao, H.; and Jiang, S. 2021.
\newblock Seeking Common but Distinguishing Difference, A Joint Aspect-based Sentiment Analysis Model.
\newblock In Moens, M.-F.; Huang, X.; Specia, L.; and Yih, S. W.-t., eds., \emph{Proceedings of the 2021 Conference on Empirical Methods in Natural Language Processing}, 3910--3922. Online and Punta Cana, Dominican Republic: Association for Computational Linguistics.

\bibitem[{Ju et~al.(2021)Ju, Zhang, Xiao, Li, Li, Zhang, and Zhou}]{ju-etal-2021-joint}
Ju, X.; Zhang, D.; Xiao, R.; Li, J.; Li, S.; Zhang, M.; and Zhou, G. 2021.
\newblock Joint Multi-modal Aspect-Sentiment Analysis with Auxiliary Cross-modal Relation Detection.
\newblock In \emph{Proceedings of the 2021 Conference on Empirical Methods in Natural Language Processing}, 4395--4405. Online and Punta Cana, Dominican Republic: Association for Computational Linguistics.

\bibitem[{Khan and Fu(2021)}]{DBLP:conf/mm/0001F21}
Khan, Z.; and Fu, Y. 2021.
\newblock Exploiting {BERT} for Multimodal Target Sentiment Classification through Input Space Translation.
\newblock In Shen, H.~T.; Zhuang, Y.; Smith, J.~R.; Yang, Y.; C{\'{e}}sar, P.; Metze, F.; and Prabhakaran, B., eds., \emph{{MM} '21: {ACM} Multimedia Conference, Virtual Event, China, October 20 - 24, 2021}, 3034--3042. {ACM}.

\bibitem[{LeCun et~al.(2006)LeCun, Chopra, Hadsell, Ranzato, and Huang}]{lecun2006tutorial}
LeCun, Y.; Chopra, S.; Hadsell, R.; Ranzato, M.; and Huang, F. 2006.
\newblock A tutorial on energy-based learning.
\newblock \emph{Predicting structured data}, 1(0).

\bibitem[{Li et~al.(2023{\natexlab{a}})Li, Li, Savarese, and Hoi}]{DBLP:journals/corr/abs-2301-12597}
Li, J.; Li, D.; Savarese, S.; and Hoi, S. C.~H. 2023{\natexlab{a}}.
\newblock {BLIP-2:} Bootstrapping Language-Image Pre-training with Frozen Image Encoders and Large Language Models.
\newblock \emph{CoRR}, abs/2301.12597.

\bibitem[{Li et~al.(2023{\natexlab{b}})Li, Zhang, Zhao, Yang, and Yang}]{DBLP:journals/corr/abs-2307-00360}
Li, Z.; Zhang, S.; Zhao, H.; Yang, Y.; and Yang, D. 2023{\natexlab{b}}.
\newblock BatGPT: {A} Bidirectional Autoregessive Talker from Generative Pre-trained Transformer.
\newblock \emph{CoRR}, abs/2307.00360.

\bibitem[{Ling, Yu, and Xia(2022)}]{ling-etal-2022-vision}
Ling, Y.; Yu, J.; and Xia, R. 2022.
\newblock Vision-Language Pre-Training for Multimodal Aspect-Based Sentiment Analysis.
\newblock In \emph{Proceedings of the 60th Annual Meeting of the Association for Computational Linguistics (Volume 1: Long Papers)}, 2149--2159. Dublin, Ireland: Association for Computational Linguistics.

\bibitem[{Liu et~al.(2019)Liu, Ott, Goyal, Du, Joshi, Chen, Levy, Lewis, Zettlemoyer, and Stoyanov}]{DBLP:journals/corr/abs-1907-11692}
Liu, Y.; Ott, M.; Goyal, N.; Du, J.; Joshi, M.; Chen, D.; Levy, O.; Lewis, M.; Zettlemoyer, L.; and Stoyanov, V. 2019.
\newblock RoBERTa: {A} Robustly Optimized {BERT} Pretraining Approach.
\newblock \emph{CoRR}, abs/1907.11692.

\bibitem[{Lu et~al.(2018)Lu, Neves, Carvalho, Zhang, and Ji}]{lu-etal-2018-visual}
Lu, D.; Neves, L.; Carvalho, V.; Zhang, N.; and Ji, H. 2018.
\newblock Visual Attention Model for Name Tagging in Multimodal Social Media.
\newblock In \emph{Proceedings of the 56th Annual Meeting of the Association for Computational Linguistics (Volume 1: Long Papers)}, 1990--1999. Melbourne, Australia: Association for Computational Linguistics.

\bibitem[{Peng, Li, and Zhao(2021)}]{DBLP:journals/corr/abs-2109-06719}
Peng, L.; Li, Z.; and Zhao, H. 2021.
\newblock Sparse Fuzzy Attention for Structured Sentiment Analysis.
\newblock \emph{CoRR}, abs/2109.06719.

\bibitem[{Peng et~al.(2023)Peng, Li, Zhang, Du, and Zhao}]{peng-etal-2023-fsuie}
Peng, T.; Li, Z.; Zhang, L.; Du, B.; and Zhao, H. 2023.
\newblock {FSUIE}: A Novel Fuzzy Span Mechanism for Universal Information Extraction.
\newblock In \emph{Proceedings of the 61st Annual Meeting of the Association for Computational Linguistics (Volume 1: Long Papers)}, 16318--16333. Toronto, Canada: Association for Computational Linguistics.

\bibitem[{Radford et~al.(2021)Radford, Kim, Hallacy, Ramesh, Goh, Agarwal, Sastry, Askell, Mishkin, Clark, Krueger, and Sutskever}]{DBLP:conf/icml/RadfordKHRGASAM21}
Radford, A.; Kim, J.~W.; Hallacy, C.; Ramesh, A.; Goh, G.; Agarwal, S.; Sastry, G.; Askell, A.; Mishkin, P.; Clark, J.; Krueger, G.; and Sutskever, I. 2021.
\newblock Learning Transferable Visual Models From Natural Language Supervision.
\newblock In Meila, M.; and Zhang, T., eds., \emph{Proceedings of the 38th International Conference on Machine Learning, {ICML} 2021, 18-24 July 2021, Virtual Event}, volume 139 of \emph{Proceedings of Machine Learning Research}, 8748--8763. {PMLR}.

\bibitem[{Sun et~al.(2021)Sun, Wang, Zhang, Su, and Weng}]{DBLP:conf/aaai/0006W0SW21}
Sun, L.; Wang, J.; Zhang, K.; Su, Y.; and Weng, F. 2021.
\newblock RpBERT: {A} Text-image Relation Propagation-based {BERT} Model for Multimodal {NER}.
\newblock In \emph{Thirty-Fifth {AAAI} Conference on Artificial Intelligence, {AAAI} 2021, Thirty-Third Conference on Innovative Applications of Artificial Intelligence, {IAAI} 2021, The Eleventh Symposium on Educational Advances in Artificial Intelligence, {EAAI} 2021, Virtual Event, February 2-9, 2021}, 13860--13868. {AAAI} Press.

\bibitem[{Wang et~al.(2022{\natexlab{a}})Wang, Cai, Jiang, Xie, Tu, and Lu}]{wang-etal-2022-named}
Wang, X.; Cai, J.; Jiang, Y.; Xie, P.; Tu, K.; and Lu, W. 2022{\natexlab{a}}.
\newblock Named Entity and Relation Extraction with Multi-Modal Retrieval.
\newblock In \emph{Findings of the Association for Computational Linguistics: EMNLP 2022}, 5925--5936. Abu Dhabi, United Arab Emirates: Association for Computational Linguistics.

\bibitem[{Wang et~al.(2022{\natexlab{b}})Wang, Gui, Jiang, Jia, Bach, Wang, Huang, and Tu}]{DBLP:conf/naacl/WangGJJBWHT22}
Wang, X.; Gui, M.; Jiang, Y.; Jia, Z.; Bach, N.; Wang, T.; Huang, Z.; and Tu, K. 2022{\natexlab{b}}.
\newblock {ITA:} Image-Text Alignments for Multi-Modal Named Entity Recognition.
\newblock In Carpuat, M.; de~Marneffe, M.; and Ru{\'{\i}}z, I. V.~M., eds., \emph{Proceedings of the 2022 Conference of the North American Chapter of the Association for Computational Linguistics: Human Language Technologies, {NAACL} 2022, Seattle, WA, United States, July 10-15, 2022}, 3176--3189. Association for Computational Linguistics.

\bibitem[{Wu et~al.(2020{\natexlab{a}})Wu, Cheng, Wang, Li, and Chi}]{DBLP:conf/nlpcc/WuCWLC20}
Wu, H.; Cheng, S.; Wang, J.; Li, S.; and Chi, L. 2020{\natexlab{a}}.
\newblock Multimodal Aspect Extraction with Region-Aware Alignment Network.
\newblock In Zhu, X.; Zhang, M.; Hong, Y.; and He, R., eds., \emph{Natural Language Processing and Chinese Computing - 9th {CCF} International Conference, {NLPCC} 2020, Zhengzhou, China, October 14-18, 2020, Proceedings, Part {I}}, volume 12430 of \emph{Lecture Notes in Computer Science}, 145--156. Springer.

\bibitem[{Wu et~al.(2020{\natexlab{b}})Wu, Zheng, Cai, Chen, Leung, and Li}]{DBLP:conf/mm/WuZCCL020}
Wu, Z.; Zheng, C.; Cai, Y.; Chen, J.; Leung, H.; and Li, Q. 2020{\natexlab{b}}.
\newblock Multimodal Representation with Embedded Visual Guiding Objects for Named Entity Recognition in Social Media Posts.
\newblock In Chen, C.~W.; Cucchiara, R.; Hua, X.; Qi, G.; Ricci, E.; Zhang, Z.; and Zimmermann, R., eds., \emph{{MM} '20: The 28th {ACM} International Conference on Multimedia, Virtual Event / Seattle, WA, USA, October 12-16, 2020}, 1038--1046. {ACM}.

\bibitem[{Xu et~al.(2020)Xu, Li, Lu, and Bing}]{xu-etal-2020-position}
Xu, L.; Li, H.; Lu, W.; and Bing, L. 2020.
\newblock Position-Aware Tagging for Aspect Sentiment Triplet Extraction.
\newblock In \emph{Proceedings of the 2020 Conference on Empirical Methods in Natural Language Processing (EMNLP)}, 2339--2349. Online: Association for Computational Linguistics.

\bibitem[{Xu, Mao, and Chen(2019)}]{DBLP:conf/aaai/XuMC19}
Xu, N.; Mao, W.; and Chen, G. 2019.
\newblock Multi-Interactive Memory Network for Aspect Based Multimodal Sentiment Analysis.
\newblock In \emph{The Thirty-Third {AAAI} Conference on Artificial Intelligence, {AAAI} 2019, The Thirty-First Innovative Applications of Artificial Intelligence Conference, {IAAI} 2019, The Ninth {AAAI} Symposium on Educational Advances in Artificial Intelligence, {EAAI} 2019, Honolulu, Hawaii, USA, January 27 - February 1, 2019}, 371--378. {AAAI} Press.

\bibitem[{Yan et~al.(2021)Yan, Dai, Ji, Qiu, and Zhang}]{yan-etal-2021-unified}
Yan, H.; Dai, J.; Ji, T.; Qiu, X.; and Zhang, Z. 2021.
\newblock A Unified Generative Framework for Aspect-based Sentiment Analysis.
\newblock In \emph{Proceedings of the 59th Annual Meeting of the Association for Computational Linguistics and the 11th International Joint Conference on Natural Language Processing (Volume 1: Long Papers)}, 2416--2429. Online: Association for Computational Linguistics.

\bibitem[{Yang, Na, and Yu(2022)}]{DBLP:journals/ipm/YangNY22}
Yang, L.; Na, J.; and Yu, J. 2022.
\newblock Cross-Modal Multitask Transformer for End-to-End Multimodal Aspect-Based Sentiment Analysis.
\newblock \emph{Inf. Process. Manag.}, 59(5): 103038.

\bibitem[{Yang et~al.(2021)Yang, Yu, Zhang, and Na}]{DBLP:conf/iconference/YangYZN21}
Yang, L.; Yu, J.; Zhang, C.; and Na, J. 2021.
\newblock Fine-Grained Sentiment Analysis of Political Tweets with Entity-Aware Multimodal Network.
\newblock In Toeppe, K.; Yan, H.; and Chu, S.~K., eds., \emph{Diversity, Divergence, Dialogue - 16th International Conference, iConference 2021, Beijing, China, March 17-31, 2021, Proceedings, Part {I}}, volume 12645 of \emph{Lecture Notes in Computer Science}, 411--420. Springer.

\bibitem[{Yu and Jiang(2019)}]{DBLP:conf/ijcai/Yu019}
Yu, J.; and Jiang, J. 2019.
\newblock Adapting {BERT} for Target-Oriented Multimodal Sentiment Classification.
\newblock In Kraus, S., ed., \emph{Proceedings of the Twenty-Eighth International Joint Conference on Artificial Intelligence, {IJCAI} 2019, Macao, China, August 10-16, 2019}, 5408--5414. ijcai.org.

\bibitem[{Yu, Jiang, and Xia(2020)}]{DBLP:journals/taslp/YuJX20}
Yu, J.; Jiang, J.; and Xia, R. 2020.
\newblock Entity-Sensitive Attention and Fusion Network for Entity-Level Multimodal Sentiment Classification.
\newblock \emph{{IEEE} {ACM} Trans. Audio Speech Lang. Process.}, 28: 429--439.

\bibitem[{Yu et~al.(2020)Yu, Jiang, Yang, and Xia}]{yu-etal-2020-improving-multimodal}
Yu, J.; Jiang, J.; Yang, L.; and Xia, R. 2020.
\newblock Improving Multimodal Named Entity Recognition via Entity Span Detection with Unified Multimodal Transformer.
\newblock In \emph{Proceedings of the 58th Annual Meeting of the Association for Computational Linguistics}, 3342--3352. Online: Association for Computational Linguistics.

\bibitem[{Zhao et~al.(2022)Zhao, Dong, Shi, Yan, Xu, and Li}]{zhao-etal-2022-entity}
Zhao, G.; Dong, G.; Shi, Y.; Yan, H.; Xu, W.; and Li, S. 2022.
\newblock Entity-level Interaction via Heterogeneous Graph for Multimodal Named Entity Recognition.
\newblock In \emph{Findings of the Association for Computational Linguistics: EMNLP 2022}, 6345--6350. Abu Dhabi, United Arab Emirates: Association for Computational Linguistics.

\bibitem[{Zhou et~al.(2023)Zhou, Guo, Liu, Yu, Zhang, and Yuan}]{zhou-etal-2023-aom}
Zhou, R.; Guo, W.; Liu, X.; Yu, S.; Zhang, Y.; and Yuan, X. 2023.
\newblock {A}o{M}: Detecting Aspect-oriented Information for Multimodal Aspect-Based Sentiment Analysis.
\newblock In \emph{Findings of the Association for Computational Linguistics: ACL 2023}, 8184--8196. Toronto, Canada: Association for Computational Linguistics.

\bibitem[{Zou, Yang, and Wu(2021)}]{zou-etal-2021-unsupervised}
Zou, H.; Yang, J.; and Wu, X. 2021.
\newblock Unsupervised Energy-based Adversarial Domain Adaptation for Cross-domain Text Classification.
\newblock In Zong, C.; Xia, F.; Li, W.; and Navigli, R., eds., \emph{Findings of the Association for Computational Linguistics: ACL-IJCNLP 2021}, 1208--1218. Online: Association for Computational Linguistics.

\end{thebibliography}

\newpage

\section{Appendix}

\subsection{A. Backbone inrtoduction\&hyper-parameters selection}

Specifically, the frozen image encoder contains 48 layers of 104-head Transformer layers with hidden size of 1664, the \textit{Prompt as Dual Query} module and the text encoder have 12 layers of 12-head Transformer layers and a hidden size of 768.
Hyper-parameters used during training are shown in table~\ref{tab:Hyper-parameters}

\begin{table}[h]
\small
\centering
\begin{tabular}{lcccc}
\toprule
\bf Training Stage&  $\lambda_{1}$ & $\lambda_{2}$ &$\lambda_{3}$ &learing rate \\
\midrule
Pretraing-Stage1 &2.0&2.0&1.0&5e-5\\
Pretraing-Stage2 &1.0&1.0&1.0&3e-5\\
Finetuning       &0.1&0.1&1.0&2e-5\\
\bottomrule
\end{tabular}
\caption{Hyper-parameters used during training}
\label{tab:Hyper-parameters}
\end{table}

\subsection{B. Data construction for pre-training}
\subsubsection{COCO dataset}
\begin{table}[h]
\small
\centering
\begin{tabular}{lp{5cm}}
\toprule
\bf Prompt&  ``Provide a description for image." \\
\midrule
\bf Description&  the label description of image \\
\midrule
\bf Text&  label description, Irrelevant description$1$, Irrelevant description$2$, Irrelevant description$3$. \\
\bottomrule
\end{tabular}
\caption{COCO pre-training data construction}
\label{tab:COCO_data}
\end{table}

For COCO data, each image has corresponding five descriptions. We randomly select one relevant description and three irrelevant descriptions to splice as the input text, the model needs to predict the descriptions that are relevant to the content of the picture under the guidance of prompt. Training data is constructed as table~\ref{tab:COCO_data}.

\subsubsection{ImageNet dataset}
\begin{table}[h]
\small
\centering
\begin{tabular}{lp{5cm}}
\toprule
\bf Prompt&  ``What does this image contains." \\
\midrule
\bf Description&  ``It's an image of \{image label\}." \\
\midrule
\bf Text&  Correct label, fake label$1$, fake label$2$ $\cdots$ fake label$9$. \\
\bottomrule
\end{tabular}
\caption{ImageNet pre-training data construction}
\label{tab:ImageNet_data}
\end{table}

For ImageNet data, which contains 1000 classes of images and corresponding labels, we select 100 of these classes to construct the pre-training data, and divide them into 10 groups equally. For the same group of data, the input text is constructed as a splice of ten class names, and the model needs to predict the entity class contained in the current image under the guidance of prompt. Training data is constructed as table~\ref{tab:ImageNet_data}

\end{document}